\documentclass{article}
\usepackage{spconf,amsmath,graphicx}
\usepackage[utf8]{inputenc}
\usepackage[english]{babel}
\usepackage{xcolor}
\usepackage{booktabs}
\usepackage{siunitx}
\usepackage{lineno}
\usepackage{float}

\ninept

\usepackage[capitalise]{cleveref}

\title{Conformers are All You Need for Visual Speech Recognition}
\name{Oscar Chang, Hank Liao, Dmitriy Serdyuk, Ankit Shah\textsuperscript{\textdagger}\thanks{\textsuperscript{\textdagger}Author was a research intern at the time of the work.}, Olivier Siohan}
\address{Google, Inc}
\begin{document}
\maketitle
\begin{abstract}
Visual speech recognition models extract visual features in a hierarchical manner. At the lower level, there is a visual front-end with a limited temporal receptive field that processes the raw pixels depicting the lips or faces. At the higher level, there is an encoder that attends to the embeddings produced by the front-end over a large temporal receptive field. Previous work has focused on improving the visual front-end of the model to extract more useful features for speech recognition. Surprisingly, our work shows that complex visual front-ends are not necessary. Instead of allocating resources to a sophisticated visual front-end, we find that a linear visual front-end paired with a larger Conformer encoder results in lower latency, more efficient memory usage, and improved WER performance. We achieve a new state-of-the-art of $12.8\%$ WER for visual speech recognition on the TED LRS3 dataset, which rivals the performance of audio-only models from just four years ago.
\end{abstract}

\section{Introduction}
\label{sec:intro}

Visual speech recognition (VSR) is the task of transcribing speech utterances into text based on visual inputs, which usually come in the form of sequences of video frames depicting the lips or face of the speaker. When paired with audio inputs, this task becomes known as audio-visual speech recognition (AVSR). Both VSR and AVSR have garnered increasing interest from the speech community \cite{chang2022robustness,braga2020end,braga2021closer,makino2019recurrent,serdyuk2022transformer}, especially in situations where the audio modality is missing or severely degraded due to background noise or overlapping speech \cite{afouras2018deep,tripathi2021end}. 

A typical VSR architecture is organized hierarchically, with a visual front-end at the lower level that extracts features from the raw pixels, and an encoder at the higher level that attends to these features to combine them with the decoder. While the encoder has a large receptive field in the temporal dimension, the visual front-end is distinguished by its limited temporal receptive field, and is often based on existing image-based neural networks from computer vision. There has been extensive prior work focused on integrating increasingly sophisticated visual front-ends into speech recognition models. Examples include residual networks \cite{afouras2018deep,shillingford2018large,shi2022learning,hsu2022single},  convolutional 3D networks \cite{shillingford2018large,ma2021end,makino2019recurrent}, VGGs \cite{braga2022best,chang2022robustness}, and vision transformers (ViT) \cite{serdyuk2022transformer}. However, these visual front-ends often incur a huge memory overhead, which has hitherto prevented the size of state-of-the-art VSR models (e.g.~$0.3$B params in \cite{serdyuk2022transformer}) from catching up to their audio-only counterparts (e.g.~$0.6$B params in \cite{chiu2022self}). The visual front-end also incurs increased latency due to the increased processing time for each individual frame. 

In this paper, we conduct a detailed profiling study in \cref{sec:profiling} to analyse the efficiency of different visual front-ends. We found that the latency of VGG and ViT front-ends does not scale well with the batch size, causing them to take up as much processing time as a Conformer-based encoder itself. This is undesirable since the front-end itself should take up a small fraction of the computational budget compared to the encoder. By replacing a ViT front-end with a linear projection (LP), we were able to fit a Conformer model that is twice as big into TPU memory. This also trains at twice the speed compared to the state-of-the-art VSR architecture in~\cite{serdyuk2022transformer}. Our work corroborates recent findings in the computer vision literature that bigger patch sizes help the memory efficiency and latency of ViTs~\cite{steiner2021train}. Our model is a continuation of that trend, since an LP front-end with a Conformer-based encoder can be interpreted as a one-patch model.

\begin{figure}[t]
    \centering
    \includegraphics[width=0.5\textwidth]{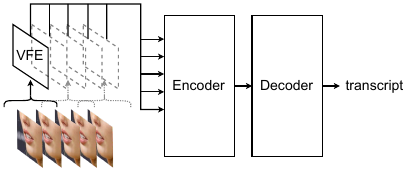}
    \caption{A typical end-to-end visual speech recognition network. At the lower level, the video is processed using a visual front-end, typically a CNN whose receptive field in the temporal dimension is limited. At the higher level, the features from the front-end are processed by an encoder, typically a Transformer-based neural network like a Conformer that attends to a wide temporal receptive field. Finally, the transcript is produced by the decoder (LSTM with the RNN-T loss).
    }
    \label{fig:lp_Conformer_diagram}
    \vspace{-4mm}
\end{figure}

The proposed one-patch model performs extremely well, achieving a new state-of-the-art word error rate of $12.8\%$ on the widely used TED LRS3 benchmark. This shows that complex front-ends are not necessary for the task of VSR, where it is more important to attend to the relationship between different video frames as opposed to the relationship of different pixels within a frame. We also conduct audio-visual diarization experiments in \cref{sec:results}, and show that our proposed method yields a $15\%$ improvement in diarization error rate compared to an equivalent model with a VGG front-end. Finally, we conduct a robustness analysis, as in \cite{chang2022robustness}, to verify that our audio-visual model degrades gracefully in scenarios of missing video.

\section{Visual Speech Recognition Architectures}
\label{sec:mod}
In this section, we introduce the visual front-end and the encoder architectures discussed in our paper.

\subsection{Visual Front-End}

\textbf{VGG} The VGG network is a convolutional network that was originally introduced for image classification in \cite{simonyan2014very}. There are two primary means of adapting a vanilla VGG for image classification, which takes 2D inputs, to VSR, which takes 3D inputs. The first is to use 3D convolutions instead of 2D ones, like in \cite{shillingford2018large,makino2019recurrent}. We denote this approach as \textbf{Conv3D}. The second is to decompose the 3D convolutions into separate 2D spatial convolutions and 1D temporal convolutions for improved memory efficiency. We use the $10$-layer (2+1)D VGG in \cite{braga2022best} for our profiling and AVSR experiments, and denote it as \textbf{VGG}. 

\noindent \textbf{ViT} Vision transformers were originally introduced for image classification in \cite{dosovitskiy2020image}. They work by slicing the input image into 2D patches, and applying a linear projection on the patches to form token embeddings. We use the $6$-layer ViT in \cite{serdyuk2022transformer} that has been adapted for VSR by using bigger 3D patches.

\noindent \textbf{LP} In our work, we introduce the embarrassingly simple Linear Projection front-end. Linear projections use a single matrix multiplication, which has been optimized to run with high computational and memory efficiency on modern accelerator hardware. To further speed up this computation, we apply the linear projection on down-sampled video frames (so there is no temporal component). In our experiments, we downsample the $128$x$128$x$3$ image to $64$x$64$x$3$ for VSR, and $32$x$32$x$3$ for AVSR.

\subsection{Encoder}
\textbf{Transformer} The Transformer was originally introduced for machine translation in \cite{vaswani2017attention}, which operated on tokenized word embeddings. It has since been adapted for speech recognition, and is typically applied on short-time Fourier transform spectrogram features instead of on the raw audio waveform. For VSR, it is applied on the visual embeddings produced by the front-end. For our audio-visual diarization experiments, we use the specific architecture introduced in the prior work of \cite{braga2022best} for a baseline comparison.

\noindent \textbf{Conformer} Conformers are convolution-augmented transformers that were initially introduced for audio-only speech recognition in \cite{gulati2020conformer}. A Conformer encoder block consists of a sequence of four sub-networks: a feed-forward network (FFN), multi-headed self attention (MHSA), a 1D convolution layer, and a second FFN. Mathematically, for input $x_i$, the output $y_i$ can be expressed as such:
\begin{equation}
    \tilde{x_i} = x_i + \frac{1}{2} \textrm{FFN}(x_i) 
\end{equation}
\begin{equation}
     {x_i}'= \tilde{x_i} + \textrm{MHSA} (\tilde{x_i})
\end{equation}
\begin{equation}
    {x_i}'' = {x_i}' + \textrm{Conv}({x_i}')
\end{equation}
\begin{equation}
    {y_i} = \textrm{LayerNorm}({x_i}'' + \frac{1}{2} \textrm{FFN}({x_i}''))
\end{equation}
Conformer encoders have been used to obtain state-of-the-art results for both VSR and AVSR \cite{serdyuk2022transformer,chang2022robustness}, but only in conjunction with a sophisticated ViT or VGG based visual front-end. In our work, we show new state-of-the-art results on the TED-LRS3 benchmark, by using Conformer encoders with an LP front-end.

\section{Profiling Analysis}
\label{sec:profiling}
In this section, we perform two kinds of profiling analysis. The first is an offline profiling that isolates a neural network component on a single TPU core. The second is an online profiling that runs an entire speech recognition network on live production machine learning infrastructure with $512$ TPU cores.

\subsection{Offline Profiling}
We first measure the absolute latency of two commonly used visual front-ends, namely the ten-layer VGG network in \cite{braga2022best} ($7.0$M params) and the six-layer ViT network in \cite{serdyuk2022transformer} ($37.3$M params). As a comparison, we also profile a $3072$x$512$ linear projection ($1.6$M params), and a fifteen-layer Conformer ($96.4$M params), which is the typical size of an encoder network. For a single TPU core, both the VGG and the ViT were unable to fit a batch size of $16$ in memory, whereas the LP and the Conformer did fit in memory. At batch size $8$, we noticed that the VGG and ViT incurred a $263.7$ms and $366.8$ms latency for a forward pass respectively, while the LP and Conformer took $25.8$ms and $288.9$ms respectively. Because we expect the visual front-end to be responsible for only a small part of the feature extraction and the encoder to do most of the heavy lifting, it is surprising how computationally and memory intensive existing visual front-ends like the VGG and ViT are compared to the Conformer encoder.

\begin{figure}[t]
    \centering
    \includegraphics[width=0.48\textwidth]{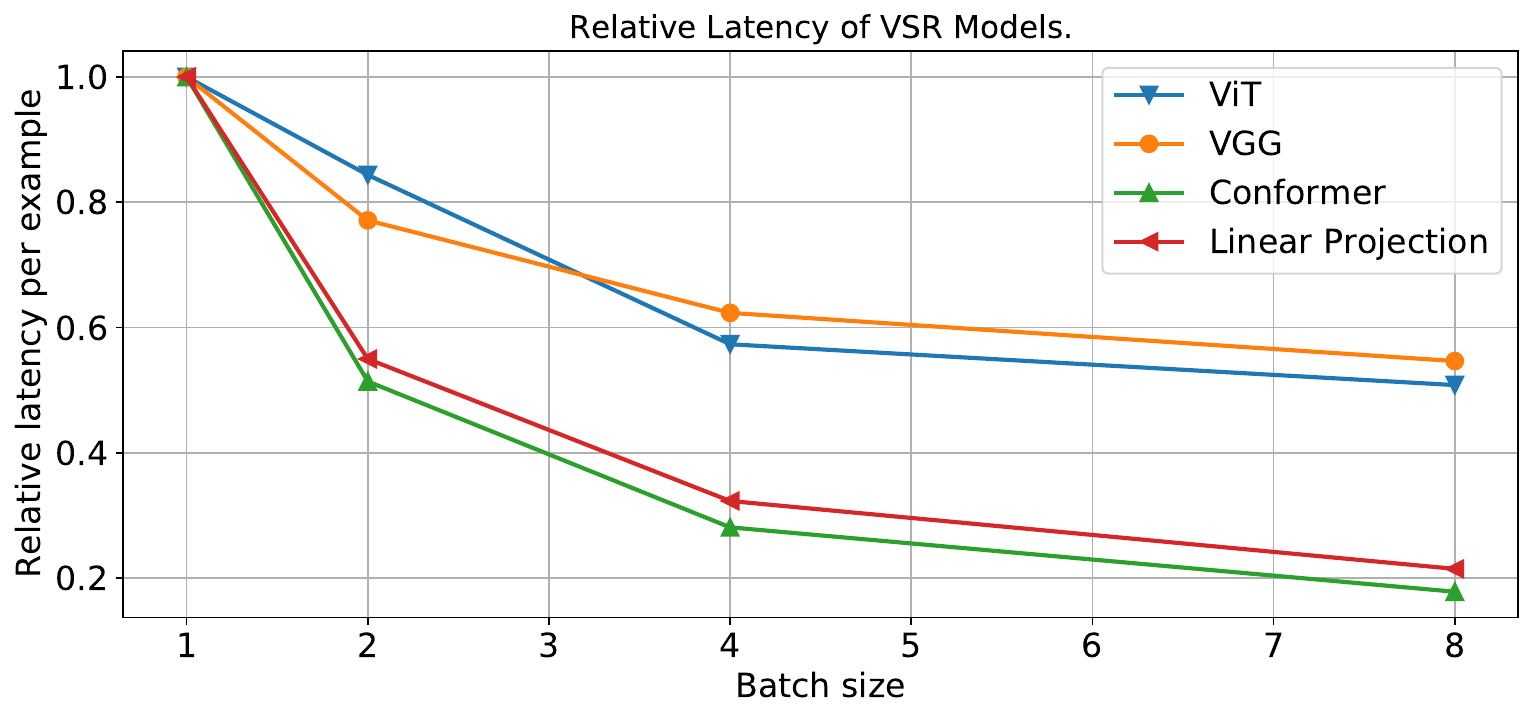}
    \caption{Effect on TPU latency upon scaling the batch size. The latency is measured as the amount of time it takes for a forward pass relative to that for a single example, i.e.~batch size $1$.}
    \label{fig:relative_tpu_latency}
\end{figure}

In~\cref{fig:relative_tpu_latency}, we plot a graph of TPU latencies for each architecture at each batch size relative to the latency for batch size~$1$, where the latency is measured as the amount of time it takes for a forward pass through the network. The shape of this graph looks similar even when varying the number of layers or the size of the LP, and thus can be relied upon to qualitatively understand the model's ability to parallelize the computation across the mini-batch. We observe that at batch size $8$, both the LP and the Conformer scale extremely well, taking about $0.2$ the latency relative to batch size $1$, unlike the ViT or VGG, which take about $0.5$ the latency.

\subsection{Online Profiling}
Next, we do an online profiling of the current state-of-the-art VSR model, the ViT Conformer \cite{serdyuk2022transformer}, which is composed of a $6$-layer ViT visual front-end and a $17$-layer Conformer encoder with model dimension $512$. We compare it with our proposed LP Conformer model, which uses an LP front-end and a $16$-layer Conformer encoder with model dimension $1024$. Both models use full-context attention layers in their Conformer encoder. To make the comparison fair, we run both using the same LSTM-based RNN-T decoder, on the maximum batch size (in powers of $2$), on the same TPU hardware, with the same XLA compiler settings.

In \cref{table:profiling}, we note that the LP Conformer trains at $4.2$k examples per second, which is more than twice as fast as the ViT Conformer. In-depth profiling reveals that the ViT Conformer is bottle-necked by reshape operations because it operates on multiple patches per frame, while the LP Conformer avoids this problem by essentially using only one big patch per frame. The LP Conformer also manages to afford twice the number of parameters ($0.57$B versus $0.31$B) for roughly the same amount of TPU memory, leading to significantly improved memory efficiency ($59.9$ bytes versus $103.2$ bytes per parameter). This huge increase in memory efficiency can be attributed to the XLA compiler being more optimized for modularized architectures (i.e.~a stack of Conformer blocks, as opposed to a mixture of Conformer and ViT blocks). Despite being twice as big, the XLA graph for the LP Conformer also compiled $29\%$ faster than that of the ViT Conformer.

\begin{table}[t]
    \centering
    \caption{Profiling comparison between ViT Conformer and LP Conformer.}
    
    \begin{center}
    \begin{footnotesize}
    \begin{sc}
    \begin{tabular}{lccc}
        \toprule
        {\bfseries Model} & {\bfseries Train Speed} & {\bfseries \#Params} & {\bfseries Memory} \\
        \midrule
        ViT Conformer~\cite{serdyuk2022transformer} & $1.9$k/s & $0.31$B & $29.79$G \\
        LP Conformer & $4.2$k/s & $0.57$B & $31.80$G \\
        \bottomrule
    \end{tabular}
    \end{sc}
    \end{footnotesize}
    \end{center}
    \vspace{-9mm}
    \label{table:profiling}
\end{table}

\section{Experimental Setup}
\subsection{Safe AI Principles}
Because of the sensitive nature of the technology used in this work, we took care to conduct our research in a safe and responsible manner by following the Google AI principles \cite{google2018artificial}.

\subsection{Data}

The visual features in all our datasets are prepared by running the MediaPipe face detector \cite{lugaresi2019mediapipe} to extract $128$x$128$ RGB mouth tracks. The acoustic features consist of $80$-dimensional log-mel filterbank coefficients extracted from a $25$ms Hann window with a stride of $10$ms; every three frames are stacked to yield a 240-dimensional frame every 30ms.  The text transcripts are tokenized at the character level. We use the following three datasets in this paper.

\noindent \textbf{TED-LRS3}: TED-LRS3 \cite{afouras2018lrs3} is the only publicly available AVSR dataset that has a non-restrictive license (Creative Commons) and is compliant with European data protection laws. As such, it has been widely used by the community as an evaluation benchmark. It contains over $400$ hours of TED and TEDx videos in segments of $6$ seconds, covering a wide range of speakers ($>5$k).

\noindent \textbf{YT}: We mine public YouTube videos for audio-visual speech with high confidence transcripts by following the same procedure as \cite{makino2019recurrent,serdyuk2022transformer}. This procedure involves extracting segments of video (islands) where the user-uploaded captions matches the result of a pre-trained audio-only speech recognition system. Then, an audio-visual synchronization classifier \cite{chung2016out,shillingford2018large} is used to filter out cases where the found face is a still image or contains a dubbed voice. The YT dataset contains $100$k hours of transcribed video segments that are up to $15$ seconds long.

\noindent \textbf{MEET360}: MEET360 is a simulated audio-visual dataset that was collected internally. Volunteers were asked to conduct $20$-minute meetings in a conference room, and recorded by a $360$-degree GoPro camera. The resultant spherical video is then projected to a hi-res $2$D video, and transcribed by human annotators. The $2$D video is segmented by the ground-truth non-speaking regions to yield clips averaging $25$ seconds, totaling $11$ hours of video. Unlike TED-LRS3 and YT where a single video track is associated with each audio track, there are multiple video tracks associated with each audio track in MEET360.

\section{Results}
\label{sec:results}
We conduct three experiments to validate our proposal of the LP Conformer. For experiments with visual-only inputs, each frame is downsampled to $64$x$64$, processed by the LP front-end, then passed to a $16$-layer full-context Conformer encoder with model dimension $1024$ (c.f.~\cref{table:profiling}). For experiments with audio-visual inputs, each frame is downsampled to $32$x$32$, then processed by the LP front-end and two Conformer layers before being concatenated to the acoustic features. The concatenated features are then passed to a $15$-layer full-context Conformer encoder with model dimension $512$. No external LM was used.

\subsection{Visual Speech Recognition}
First, we investigate the performance of the LP Conformer on the VSR task, where text transcripts are produced solely from visual inputs. We follow the experimental setup in \cite{makino2019recurrent,serdyuk2022transformer} by training on YT, and fine-tuning on TED-LRS3. For the decoder, we use RNN-T, a $9$-layer LSTM with cell size $2048$, embedding dimension $128$, and beam width $8$. The model was trained for $300$k steps using a global batch size of $16,384$ and the Adam optimizer with a $3$k-step warmup and a peak learning rate of $5$e-$4$ that is cosine-annealed to $5$e-$5$.

Our results are reported in \cref{table:vo_wer} with comparisons to prior work. We significantly outperform prior work, achieving $12.8\%$ WER, which is a $25\%$ improvement over the previous state-of-the-art ViT Conformer at $17.0\%$ WER. This result convincingly demonstrates that sophisticated visual front-ends like residual networks, 3D convolutional neural networks, or vision transformers are not necessary for visual speech recognition. Moreover, our video-only model rivals the performance of audio-only models from just four years ago, for example the CTC and seq2seq models from \cite{afouras2018deep} that performed at $13.8\%$ and $9.0\%$ respectively. That same audio-only model performed at $65.6\%$ WER when tested under noisy acoustic conditions, whereas our video-only model can sustain its performance under those conditions. 

We hypothesize that the performance gains come primarily from the increased training efficiency of the LP Conformer, which allows it to take full advantage of the $100$k hours of training data in YT. To test our hypothesis, we reproduced the ViT Conformer in \cite{serdyuk2022transformer}, but trained it for a total of $18$ days and found that it was able to achieve $13.2\%$ WER, which is comparable to the $12.8\%$ result with the LP Conformer with only $8$ days of training. This is reminiscent of early representation learning results like word2vec \cite{mikolov2013efficient}, which showed that given a fixed compute budget, a one-layer MLP trained on a huge amount of data can actually out-perform large neural networks trained on a smaller amount of data.

\begin{table*}[t]
    \centering
    \caption{Comparison against prior work on TED LRS3. VO denotes video-only, AV denotes audio-visual.}
    \begin{center}
    \begin{small}
    \begin{sc}
    \begin{tabular}{lccccc}
        \toprule
        {\bfseries Model} & {\bfseries Visual Front-End} & {\bfseries Encoder}      & {\bfseries Train Data (Hours)} & {\bfseries VO-WER} & {\bfseries AV-WER} \\
        \midrule
        TM-CTC~\cite{afouras2018deep} & ResNet & Transformer & $\;\,0.5$k & $74.7$ & $12.3$ \\
        TM-seq2seq~\cite{afouras2018deep} & ResNet & Transformer & $\;\,0.5$k & $59.9$ & $8.0$ \\
        V2P~\cite{shillingford2018large} & Conv3D & LSTM & $\;\,4$k & $55.1$ & $-$\\
        ResNet Conformer~\cite{ma2021end} & Conv3D+ResNet & Conformer & $\;\,0.5$k & $43.3$ & $2.3$ \\
        RNN-T~\cite{makino2019recurrent} & Conv3D & LSTM & $31$k & $33.6$ & $4.5$\\
        VGG Transformer \cite{braga2022best} & VGG & Transformer & $90$k & - & $3.0$ \\
        u-HuBERT~\cite{hsu2022single} & ResNet & Transformer & $\;\,2$k & $27.2$ & $1.2$ \\
        AV-HuBERT~\cite{shi2022learning} & ResNet & Transformer & $\;\,2$k & $26.9$ & $1.3$ \\
        ViT 3D \cite{serdyuk2022transformer} & ViT & Conformer & $90$k & $17.0$ & $1.6$ \\
        VGG Conformer \cite{chang2022robustness} & VGG & Conformer & $100$k & - & $0.9$ \\
        \midrule
        LP Conformer (Ours) & Linear Projection & Conformer & $100$k & $\boldsymbol{12.8}$ & $\boldsymbol{0.9}$ \\
        \bottomrule
    \end{tabular}
    \end{sc}
    \end{small}
    \end{center}
    \vspace{-8mm}
    \label{table:vo_wer}
\end{table*}

\subsection{Audio-Visual Diarization}
Next, we study audio-visual diarization, where the task is to determine the speaker identity associated with each segment of audio, i.e.~\emph{who spoke when}, by using the visual information available. We use diarization error rate (DER) and word diarization error rate (WDER) to measure diarization performance. DER is the percentage duration of audio that is labeled with the wrong speaker, for frames of audio that are labeled with a speaker.
\begin{equation}
\textrm{DER} =
\frac{
\textrm{false alarm} +
\textrm{missed detection} +
\textrm{confusion}}
{\textrm{total}},
\end{equation}
Often, the DER does not match the perceived speaker diarization performance due to the measurement unit being time rather than words. Thus, an alternate metric to consider is WDER, which can be computed as the percentage of words that are mislabeled with the wrong speaker,
\begin{equation}
\textrm{WDER} =
\frac{
\textrm{C}_\textrm{IS} +
\textrm{S}_\textrm{IS} +
\textrm{I}_\textrm{IS}}
{\textrm{\textrm{C}+\textrm{S}+\textrm{I}}},
\end{equation}
where $\textrm{C}_\textrm{IS}$, $\textrm{S}_\textrm{IS}$, and $\textrm{I}_\textrm{IS}$ are the correct, substituted, and inserted words with incorrect speaker labeling. $C+S+I$ is the total number of hypothesis words. Unlike DER, WDER can only be measured when the diarization system outputs a transcript.

Like \cite{braga2020end,braga2021closer,braga2022best}, we equip an existing speech recognition model with diarization capabilities by incorporating an attention mechanism for face selection. We simulate conditions of multiple faces on the YT dataset by re-using video tracks from other examples in the batch during training, and test our model on the MEET360 dataset. In \cref{table:meet360}, we compare the LP Conformer against the VGG Transformer in \cite{braga2022best} (VGG front-end with a Transformer encoder) and the VGG Conformer in \cite{chang2022robustness} (VGG front-end with a Conformer encoder). We found that the LP Conformer performed the best for both speech recognition and diarization, with the LP Conformer achieving a big $26\%$ relative DER improvement and $32\%$ relative WDER improvement over the VGG Transformer baseline. We also found that the LP Conformer trained about $20\%$ faster than the VGG Conformer, while using less TPU memory to do so.

\begin{table}[t]
    \centering
    \caption{Word error rate (WER), diarization error rate (DER), and word diarization error rate (WDER) of audio-visual diarization models on MEET360.}
    \begin{center}
    \begin{small}
    \begin{sc}
    \begin{tabular}{lccc}
        \toprule
        {\bfseries Model} & {\bfseries WER} & {\bfseries DER} & {\bfseries WDER}\\
        \midrule
        VGG Transformer~\cite{braga2022best} & $25.7$ & $21.1$ & $17.2$ \\
        VGG Conformer~\cite{chang2022robustness} & $24.8$ & $18.4$ & $13.8$ \\
        LP Conformer & $\boldsymbol{24.5}$ & $\boldsymbol{15.6}$ & $\boldsymbol{11.7}$ \\
        \bottomrule
    \end{tabular}
    \end{sc}
    \end{small}
    \end{center}
    \vspace{-8mm}
    \label{table:meet360}
\end{table}

\subsection{Robustness to Missing Video}
Robustness to missing video is an essential feature of AVSR systems, because speakers can often be occluded or move off-screen \cite{chang2022robustness}. We conducted a robustness analysis following~\cite{chang2022robustness} to ensure that AVSR systems built using the LP Conformer are robust to missing video.

For test suite $\mathcal{T}$, train-time robustness is achieved when a model trained with audio-visual inputs outperforms itself when trained with audio-only inputs, i.e.~$\forall T \in \mathcal{T}, \textit{WER}(\mathcal{M}_\text{AV}) \leq \textit{WER}(\mathcal{M}_\text{AO}, T)$. Test-time robustness is achieved when an audio-visual model performs better when more video information is available at test time, i.e.~$\forall T_i,T_j \in \mathcal{T}, T_i \leq T_j \implies \textit{WER}(\mathcal{M}_\text{AV}, T_i) \leq \textit{WER}(\mathcal{M}_\text{AV}, T_j)$. A model is considered \emph{robust to missing video} with respect to a given test suite if it achieves both train-time and test-time robustness. We use the same missing video test suites as in~\cite{chang2022robustness}: \emph{Drop Utterance} (each video dropped with Bernoulli probabilities), \emph{Drop Frame} (each frame dropped with Bernoulli probabilities), \emph{Drop Start} (contiguous frames dropped from the start), \emph{Drop Middle} (contiguous frames dropped from the middle), and \emph{Drop End} (contiguous frames dropped from the end).

\begin{figure}[t]
    \centering
    \includegraphics[width=0.48\textwidth,height=3.74cm]{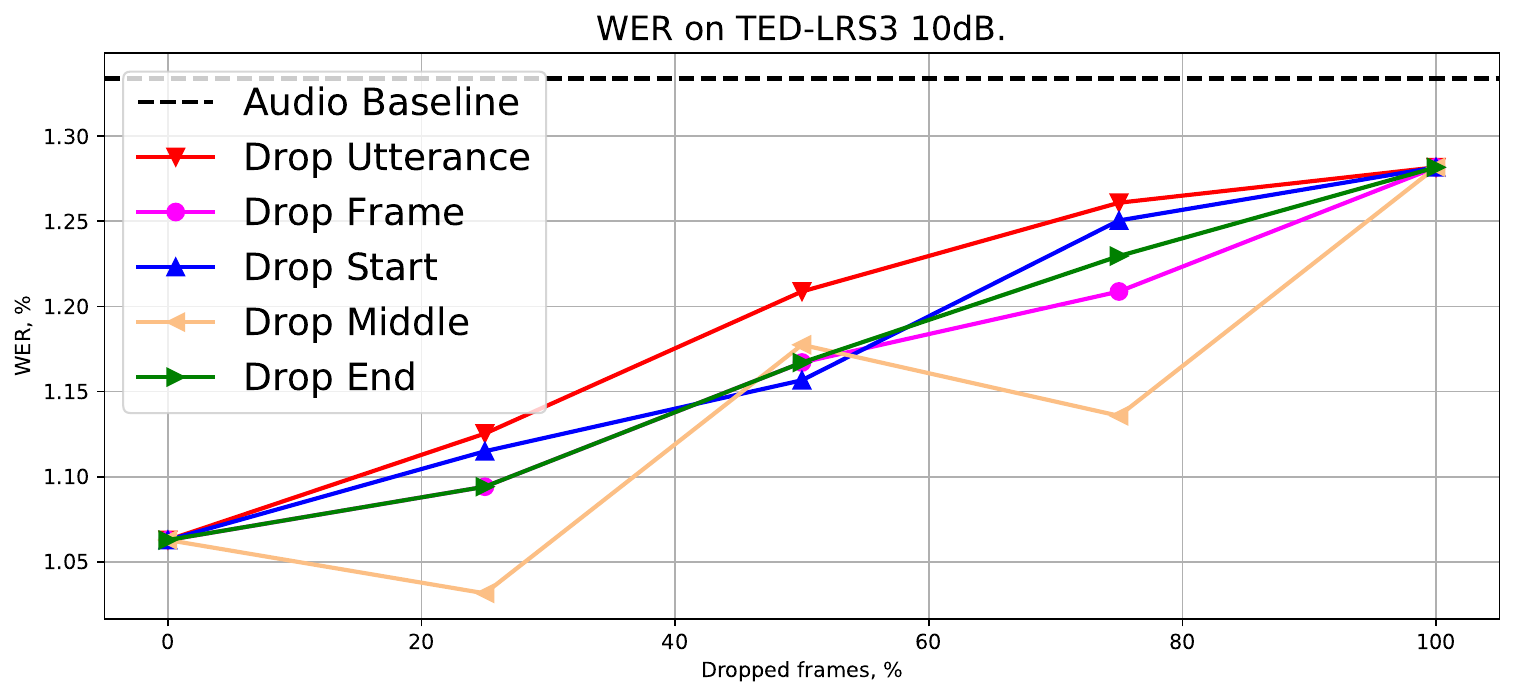}
    \caption{Robustness to missing video analysis. We evaluate on TED-LRS3 corrupted by 10dB noise. We test robustness by dropping a) the whole utterance, b) individual frames, c) frames from the start of the utterance, d) frames from the middle, e) frames from the end. WERs increase monotonically (within confidence interval bounds) as the number of frames dropped increases.}
    \label{fig:robustness}
    \vspace{-4mm}
\end{figure}

We train the LP Conformer on YT by randomly dropping the video with $50\%$ probability like in \cite{chang2022robustness,hsu2022single}, and fine-tune it on TED-LRS3. Like the VGG Conformer, the LP Conformer was found to be robust to all the missing video test suites. This is exemplied in \cref{fig:robustness}, where we can see that performance deteriorates as the percentage of frames dropped increases, but is never worse than that of the equivalent audio-only model. The LP Conformer obtains state-of-the-art results on AVSR for TED-LRS3, achieving $0.9\%$ WER when trained and tested on the full audio-visual input (c.f.~\cref{table:vo_wer}).


\section{Conclusion}
Our paper showed the surprising finding that complex visual front-ends are not necessary for VSR, achieving state-of-the-art results on TED-LRS3 using a Conformer model with a simple LP front-end. We also showed for the first time on a well-studied benchmark dataset that it is possible to do lip-reading well enough that it is in the ballpark of good audio models from four years ago. A priori, it is not obvious that this is possible, since even the best human lip-readers do not perform remotely as well as the average listener. 
\vfill\pagebreak

\clearpage

\bibliographystyle{IEEEbib}
\bibliography{references}

\end{document}